\documentclass[sigconf,nonacm]{acmart}

\usepackage{amsmath,xfrac}
\usepackage{caption}
\usepackage{subcaption}
\usepackage[vskip=1em,font=itshape,leftmargin=2em,rightmargin=2em]{quoting}
\usepackage{siunitx}
\DeclareMathOperator{\E}{\mathbb{E}}

\usepackage{hyperref}
\hypersetup{
    colorlinks=true,
    linkcolor=blue,
    filecolor=magenta,      
    urlcolor=cyan,
}
\begin{document}

\title{Fairness On The Ground: Applying Algorithmic Fairness Approaches To  Production Systems}


\author{ Chloé Bakalar, Renata Barreto, Stevie Bergman, Miranda Bogen, Bobbie Chern, Sam Corbett-Davies, Melissa Hall, Isabel Kloumann, Michelle Lam, Joaquin Quiñonero Candela, Manish Raghavan, Joshua Simons, Jonathan Tannen, Edmund Tong, Kate Vredenburgh, Jiejing Zhao}
\affiliation{Facebook}


\renewcommand{\shortauthors}{Bakalar et al.}

\newcommand{\scd}[1]{{\color{red}SCD: #1}}
\newcommand{\mr}[1]{{\color{blue}MR: #1}}
\newcommand{\et}[1]{{\color{orange}ET: #1}}
\newcommand{\mh}[1]{{\color{magenta}MH: #1}}
\newcommand{\bc}[1]{{\color{purple}BC: #1}}

\begin{abstract}
 
Many technical approaches have been proposed for ensuring that decisions made by machine learning systems are fair, but few of these proposals have been stress-tested in real-world systems.
This paper presents an example of one team’s approach to the challenge of applying algorithmic fairness approaches to complex production systems within the context of a large technology company.
We discuss how we disentangle normative questions of product and policy design (like, ``how should the system trade off between different stakeholders' interests and needs?'') from empirical questions of system implementation (like, ``is the system achieving the desired tradeoff in practice?''). We also present an approach for answering questions of the latter sort, which allows us to measure how machine learning systems \emph{and} human labelers are making these tradeoffs across different relevant groups. We hope our experience integrating fairness tools and approaches into large-scale and complex production systems will be useful to other practitioners facing similar challenges, and illuminating to academics and researchers looking to better address the needs of practitioners.
 
\end{abstract}

\maketitle

\section{Introduction}

In recent years many technical approaches have been proposed for ensuring that decisions made by machine learning systems are fair. To date, however, few of these proposals have been stress-tested in real-world systems. A viable approach to achieving fairness in practice requires overcoming a number of challenges that do not similarly constrain theoretical work. 

First, as researchers increasingly acknowledge~\cite{selbst2019fairness, passi2019problem}, purely technical or statistical approaches to fairness leave unanswered important questions related to ethics and policy. An approach to fairness in practice must have means of clearly surfacing and resolving these tensions, which are not reducible to empirical questions.

Second, it has been identified that some common statistical notions of fairness may lead to unintentional and potentially harmful consequences \cite{corbett2018measure, liu2018delayed, hu2020fair}, especially to marginalized groups. This could be even more important to consider in decision systems that may affect millions or even billions of people. Instead, what is needed as a baseline is a statistical approach that makes the costs and benefits of decisions---and who these fall on---explicit. The normative decision about who the costs and benefits \emph{should} fall on then becomes part of discussions of ethics and values, or where relevant, law or policy. By disentangling statistical questions from normative ones, this approach is flexible enough to be applied to a wide range of different systems while ensuring normative decisions are made explicitly, with input from relevant stakeholders and domain experts.

Finally, technical approaches to fairness in machine learning often rely on assumptions that may not be reasonable in practice. For example, popular statistical measures of fairness like equality of odds \cite{hardt2016equality} and calibration \cite{kleinberg2016inherent} assume that measured labels constitute ground truth, which is not always the case. Methods that allow for inaccuracies in measured labels, on the other hand, often assume the true prevalence of labels among groups~\cite{johndrow2019algorithm, feldman2015certifying}, or the causal process that generated them~\cite{kilbertus2017avoiding, loftus2018causal, kusner2017counterfactual, nabi2018fair}. These assumptions are challenged when studying human reviewers and machine learning systems in dynamic environments where adversarial actors may try to evade systems intended to constrain them.

This paper presents an example of one team's approach to address these challenges within the context of a large technology company. We are not the first to identify any of these challenges, nor are our proposed solutions---taken individually---entirely novel. However, we believe that our experience integrating fairness tools and approaches into large-scale and complex production systems may be useful to other practitioners facing similar constraints, and illuminating to academics and researchers looking to better address the needs of practitioners. 

We begin in Section \ref{sec:related_work} by surveying the fairness approaches that have seen practical implementation and the challenges preventing other approaches from being widely adopted. Section \ref{sec:ppfairness} introduces our holistic approach to assessing fairness in large-scale production systems, including a discussion of how we disentangle related yet distinct notions of fairness to facilitate more actionable analysis and decision-making for non-experts. For example, we explicitly consider and investigate fairness across multiple dimensions: at the \emph{product} level, the \emph{policy} level, and in a product or system's \emph{implementation}. 
Section \ref{sec:an_approach} details more specifically the primary measurements we use to examine implementation fairness, and why we use, as a baseline, a measurement approach grounded in making the weighing of potential benefits and harms explicit rather than deploying methods where such values decisions are implied and obscured in technical choices.

Section \ref{sec:implementation_fairness} explains how this general approach to implementation fairness is applied to two different types of decisions: those made by machine learning models (Section \ref{sec:algorithmic_decisions}) and those made by human labelers (Section \ref{sec:label_bias}). In the case of machine learning predictions, we determine the system's prioritization of errors (false positives and false negatives) for different groups by measuring the \emph{prevalence at the threshold}. To study the implied prioritization of human labelers we use Signal Detection Theory (SDT) \cite{green1966signal}, an approach with a long history in psychology, to infer the latent ``thresholds'' that labelers are applying to labeling tasks affecting different groups. We conclude in Section \ref{sec:practical_challenges} with a discussion of challenges and open technical questions that complicate the application of theoretical recommendations in practice.

The following presents a lens, and collection of approaches, by which fairness can be considered in a variety of practical cases, and we emphasize the importance of context and subject matter expertise in determining the approach(es) best suited to address potential fairness-related issues in any particular product or domain.

\section{Related work}
\label{sec:related_work}

\subsection{Fairness in practice}
\label{sec:fairness_in_practice}

Despite the explosion of academic interest in methods for developing fair algorithms, fewer methods have been implemented in production machine learning systems used by governments or private companies (at least, few of these entities have been willing to publicly share their fairness approaches, if they exist).\footnote{See \citet{holstein2019improving} for a deeper investigation into the needs of machine learning practitioners that aren't being met by current methods.}

The fairness-enhancing approaches that have achieved the most practical success seem to be efforts to improve performance by adding training data, especially for underrepresented groups. For example, after \citet{buolamwini2018gender} discovered that IBM's facial gender classification system was performing poorly for dark-skinned people---and dark-skinned women in particular---IBM responded with a system trained on more representative data which reportedly reduced the error rates on dark-skinned women almost tenfold~\cite{puri2018mitigating}.\footnote{ Subsequent work on face recognition and classification systems has focused on the ethical problems inherent in these tasks, even when the tasks are performed with accuracy for many groups~\cite{raji2020saving}. This neatly illustrates the distinction between fairness in \emph{implementation} (which concerns the performance of the system for different groups), and fairness in \emph{product} and \emph{policy} design. We return to these different ``levels'' of fairness analysis in Section \ref{sec:ppfairness}.
} Similarly, when researchers at Jigsaw noticed their comment toxicity classifier was labeling innocuous comments containing identity terms (eg ``gay'' or ``muslim'') as toxic, they augmented the training data to contain more neutral phrases with these identity terms, improving the system overall \cite{dixon2018measuring}.

In the healthcare context, researchers found that training an algorithm to predict the costs a patient would incur as a proxy for healthcare need perpetuated bias against African-American patients, who tended to incur lower costs than their white counterparts, conditioned on the same level of health need~\cite{obermeyer2019dissecting}.\footnote{Among the many competing measures of fairness, the researchers chose calibration as the measure ``most relevant to the real-world use of the algorithm''. We similarly found calibration-based approaches most useful in our applications.} In response, the healthcare provider is re-evaluating its prediction practices~\cite{ledford2019millions}.
In these cases, the solution to unfairness was simply better machine learning; a concern for fairness motivated the investigations, but the resulting changes may have been adopted even if the decision makers were concerned solely with efficiency. The oft-discussed~\cite{kleinberg2016inherent, corbett2017algorithmic} tradeoff between fairness and efficiency did not apply.

Other efforts advance to goals of fairness, accountability, and transparency in practice are more procedural in nature. Proposals like ``Datasheets for Datasets'' \cite{gebru2018datasheets} and ``Model Cards'' \cite{mitchell2019model} seek to make explicit the limitations of ML systems without making prescriptions or recommendations as to how to resolve difficult policy questions. Several companies have also released toolkits to help measure multiple fairness-oriented metrics without making prescriptions about the appropriate metric to use~\cite{bellamy2018ibm,bird2020fairlearn,wexler2020whatif}. Recognizing that documentation and metrics on their own will not necessarily lead to meaningful fairness improvements, others have proposed a structured frameworks for internal algorithmic auditing informed by organizational values \cite{raji2020closing}. 

Few fairness-minded interventions that aren't purely efficiency-oriented have been publicly discussed, with some notable exceptions. In 2019, researchers at LinkedIn published details of a system used in production to explicitly gender-balance the results returned when a recruiter searches for candidates~\cite{geyik2019fairness}.
Similarly, some vendors of algorithmic hiring assessments attempt to ensure that the data-driven models they sell don't produce outcome disparities with respect to protected characteristics like race and gender, though this may be due in part to legal considerations as opposed to purely ethical ones~\cite{bogen2018help,raghavan2020mitigating,sanchez2020does}.
Practitioners at Google describe the implementation of a particular fairness metric in production, though they aren't specific about the exact setting in which they are working~\cite{beutel2019putting}.

\subsection{Bias in machine learning predictions}

The problem of bias in supervised machine learning models is likely the most studied problem in algorithmic fairness. Numerous fairness criteria have been proposed \cite{mitchell2018prediction, berk2018fairness, verma2018fairness}, along with means of learning predictors that satisfy a given criterion.

Despite this plethora of options, few approaches appear to have been implemented in consequential production systems (see \ref{sec:fairness_in_practice} for a discussion of approaches that have seen practical implementations). We suspect there are two reasons for this. First, as \citet{corbett2018measure} discuss, many fairness criteria, including those that require equal positive classification rates (demographic parity) or equal false positive/negative rates (equality of opportunity \citep{hardt2016equality}), may fail to anticipate all implications to the well-being of the people affected by a model, and can thus inadvertently \emph{harm} marginalized groups. For example, \citet{liu2018delayed} find that, in certain lending situations, equalizing false negative rates by borrowers' race would lead to predictable decreases in the credit scores of certain African American borrowers. Similarly, \citet{hu2020fair} apply tools from welfare economics to find that ``applying more strict fairness criteria that are codified as parity constraints can worsen welfare outcomes for both groups.''

Second, as \citet{kleinberg2018algorithmic} note: ``a preference for fairness should not change the choice of estimator''. In other words, it is inappropriate to change a system's \emph{predictions} to achieve any fairness goal, since this will inevitably hurt the usefulness of the predictions for all groups. Instead, a desire for fairness should change how the predictions are used to make decisions. Practitioners, acutely aware of the cost of mistakes in their domain, are less likely to choose fairness solutions that increase these costs unnecessarily.

\subsection{Bias in training labels}

Compared to bias in predictions, bias affecting the \emph{labels} used in machine learning has received less attention in the algorithmic fairness literature.
Many algorithmic fairness metrics (including popular ones like calibration \cite{corbett2017algorithmic, kleinberg2016inherent} and equality of false positive rates \cite{hardt2016equality}) make reference to the ``true'' labels in an evaluation set, implicitly assuming these labels faithfully represent the desired target of prediction.\footnote{
\citet{jacobs2019measurement} note that the labels chosen to measure unobservable theoretical constructs should themselves be examined, but we place the choice of label beyond the scope of label bias for the purpose of this paper.}
It is common for papers discussing these methods to acknowledge that the labels might be biased (some are even motivated in part by the possibility of label bias), but it's rare for a paper to measure this bias.

Some have proposed methods to remedy possible biases in labels \cite{johndrow2019algorithm, feldman2015certifying, fogliato2020fairness}. However, without measurable ground truth to fall back upon, these approaches are left making assumptions about ground truth distributions for different groups that may not be applicable in all cases. A different line of work attempts to identify the causal paths that may lead to label biases, so that the effect of these paths can be nullified \cite{kilbertus2017avoiding, loftus2018causal, kusner2017counterfactual, nabi2018fair}. Unfortunately, these approaches are very sensitive to the structure of the causal model used, which in most cases cannot be empirically verified.

Beyond this recent computer science research there is a rich literature in economics \cite{becker2010economics, knowles2001racial, arnold2018racial}, statistics \cite{simoiu2017problem, pierson2018fast}, and psychology \cite{green1966signal, mumpower2014signal} (among other fields) developing methods to measure biases in human behavior. These methods offer hope that we might be able to identify label bias in datasets based on human decisions (for example in hiring, policing, college admissions, natural language and visual classification, etc). In section \ref{sec:label_bias}, we detail how a method from psychology, Signal Detection Theory, can be adapted to detect potential bias in human-provided labels when a source of ground truth is available.

\section{A Holistic Approach to Fairness}
\label{sec:ppfairness}

Because there is no single, agreed-upon definition of fairness, operationalizing fairness at the scale of a large and federated organization requires building a shared language for describing fairness risks and conducting standard analyses, cultivating a broad understanding of values and objectives, and establishing tools, processes, and frameworks to enable teams to make informed decisions about fairness across different contexts~\cite{friedman1996value,nissenbaum2001computer}. 

These resources cannot just be technical: imposing statistical definitions of fairness on individual machine learning models by fiat without sensitivity to wider systems and contexts in which they are embedded can backfire, failing to benefit disadvantaged groups and undermining rather than promoting fairness over time. Particularly in dynamic, complex systems like online platforms, individual machine learning models may be the wrong level at which to impose substantive requirements of fairness, so a more holistic approach is required.

We thus consider fairness at three distinct but related levels: \emph{product}, \emph{policy}, and \emph{implementation}. Fairness at the product level relates to normative and descriptive questions about the product, such as:
``Are the goals of this product consistent with providing people with fair value and treating them fairly?''
and
``How should the product trade off between different stakeholders' interests and needs?''.
Fairness at the policy level considers how the values of the organization building a system are translated into rules, leading to questions like: ``does a policy prohibiting certain types of behavior within the product adequately address the unique experiences of some subpopulations?''.
Fairness at the implementation level deals with the empirical performance of the system, answering questions like:
``Are human labelers executing the policy or labelling instructions correctly?''
and 
``Are predictive models achieving the desired tradeoff between different types of errors for all subpopulations?''.
Notice that implementation fairness questions are not limited to machine learning models---we also study the human decisions used to train the models or directly intervene in the system.

These levels are of course deeply intertwined, but by separating and differentiating between them, we aim to direct analysis toward the most salient components of a system and determine whether and what changes are needed, while ensuring each component is appropriately considered in relation to the others. For example, in order to determine the relevant models within a product to assess for implementation concerns, practitioners must understand the intended goal, structure, and use of the product as well as any policies or similar rules that may have shaped or constrained the model’s training data. If models are analyzed and found to have no implementation disparities across subgroups, but fairness concerns remain, practitioners then know to focus more deeply on the product design and policies in order to assess whether they appropriately consider the needs and harms of groups that may be impacted by that product. In other words, it is important to consider not only whether rules are being applied appropriately to all, but whether the rules themselves, or the structure in which they are situated, are fair, just and reasonable. 

Unfairness stemming from a narrowly drafted policy can be escalated to and remedied by policy stakeholders, while unfairness resulting from poor machine learning implementation can be referred to machine learning engineers for remediation. Unfairness at the product level, meanwhile, may require a fundamental reimagination of a product’s goals and objectives, requiring significant reallocation of resources by product leadership. By disentangling these three layers, unarticulated tradeoffs can be more explicitly enumerated, and disagreements about those tradeoffs can be situated with the relevant organizational decision-making frameworks.

We have found that disentangling fairness into component dimensions in this way more constructively facilitates conversations among those creating systems who are less familiar with the rich array of potential fairness-related harms and methodologies to address them, since technical work to investigate implementation concerns can be appropriately situated in qualitative conversations related to policies, legal obligations, user and stakeholder expectations, and real-world harm.

\subsection{Fairness as a process}
\label{sec:fairness_as_a_process}

Fairness is an essentially contested concept \cite{dworkin2002sovereign}, and significant interdisciplinary and intradisciplinary differences exist regarding how fairness is approached and evaluated. Scholars and practitioners from computer science, law, and philosophy, for example, may see fairness in very different lights, while related policy and regulatory notions may evolve over time.
As such, there is often considerable disagreement about what fairness entails overall, let alone at a product-specific level. It is unrealistic to presume, and would be irresponsible to claim, that simply deploying tools, checklists, or frameworks is sufficient to fully mitigate fairness risks. 

Indeed, a large proportion of fairness risks require weighing difficult tradeoffs, including seeking input from subject matter experts and people with lived experiences related to the potential harm, to inform the ultimate consensus that shapes how and to what extent fairness risks can be mitigated. Such deliberations require actionable ethical frameworks and qualitative research to fill in the gaps left by quantitative fairness approaches. 

These processes must also be iterative in order to account for both evolving notions of fairness and justice, to allow for increasing degrees of sophistication in analyses, and to account for systems and products that themselves evolve over time. For example, subsequent analysis might broaden beyond a targeted assessment of an individual model to consider the more complex fairness questions that arise in compound or dynamic systems where multiple models may interact \cite{dwork2020pipelines, hu2018short}, or ongoing monitoring may reveal that the cumulative effects of previously deployed fairness interventions have over time introduced unintended harms of their own, requiring the reevaluation of prior decisions about effective remedies to individual model-level unfairness.

Importantly, an iterative process acknowledges that fairness is never fully ``solved,'' but rather encourages ongoing consideration of fairness throughout the product development lifecycle while preserving the ability to reassess what lens (or lenses) of fairness ought to apply in a particular context---and thus what method of assessment would be best suited to test for potential unfairness in that dimension. Creating space for such flexibility is an especially common need in areas where technology has illuminated a new, augmented, or resurgent fairness risk for which acceptable and expected remedies have yet to be defined, or for which consensus does not yet exist.

While an indispensable component of any holistic approach to fairness, we leave a detailed discussion of the opportunities and challenges of implementing such processes, as well as lessons learned from efforts to integrate them into organizational processes, to future work in order to discuss with sufficient degree of detail the challenges of addressing fairness at the implementation level in the context of a large and complex organization.

\section{Fairness in implementation}
\label{sec:an_approach}
At its highest level of abstraction, our approach contains three steps:
\begin{enumerate}
    \item Catalog the costs and benefits the system may produce for people from different subgroups, and how these may trade off against one another.
    \item Make explicit choices about where the system should situate itself in this space of tradeoffs. Which costs and benefits should the system prioritize, and for whom? These normative decisions should be made at the product and policy levels of the fairness analysis.
    \item Ensure the system is minimizing costs and maximizing benefits according to the chosen tradeoffs in practice. This is the goal of the implementation level of the fairness analysis, which we address in detail in this section.
\end{enumerate}
This approach most closely resembles one advocated for by \citet{mullainathan2018algorithmic}, who argues that fairness analyses should proceed from a ``description of a global welfare function.'' \citet{kasy2020fairness} also study fairness in the context of costs and benefits; their approach to identifying the prioritization of groups (which they call the groups' ``power'') implied by observed decisions closely resembles what we discuss in Section~\ref{sec:implied}. 

While we refer to "costs" in the following sections, we use this term capaciously to describe not just economic costs but broader impacts of classification error, recognizing that such quantification requires a holistic understanding of potential harms to ensure they are sufficiently captured when weighing tradeoffs. 

\begin{figure*}[t]
    \centering
    \begin{subfigure}[b]{0.49\textwidth}
         \centering
         \includegraphics[width=0.9\textwidth]{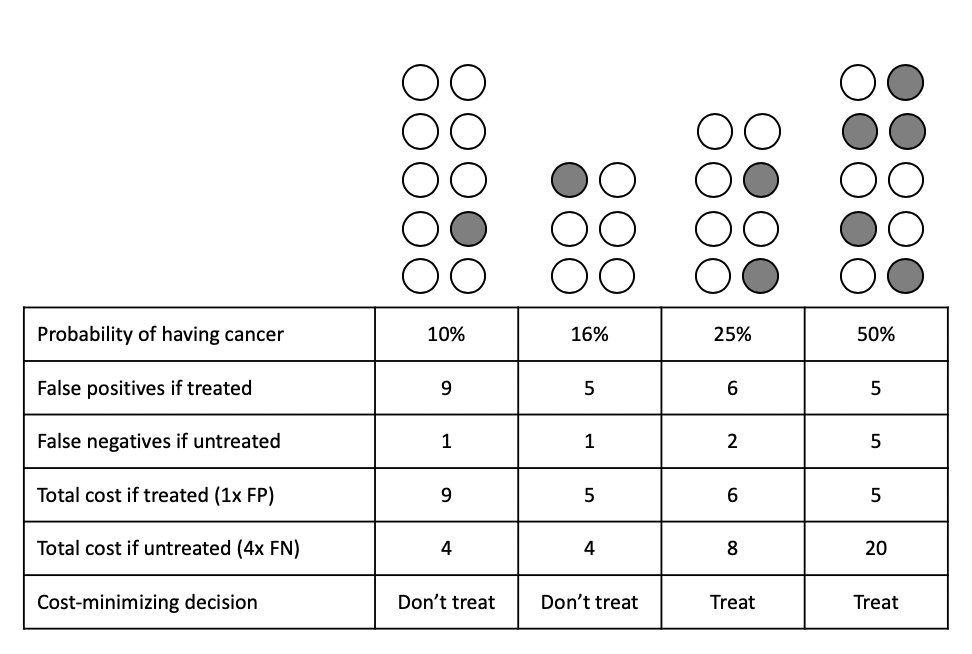}
         \caption{Female patients}
         \label{fig:cancer_risk_a}
     \end{subfigure}
     \hfill
     \begin{subfigure}[b]{0.49\textwidth}
         \centering
         \includegraphics[width=0.9\textwidth]{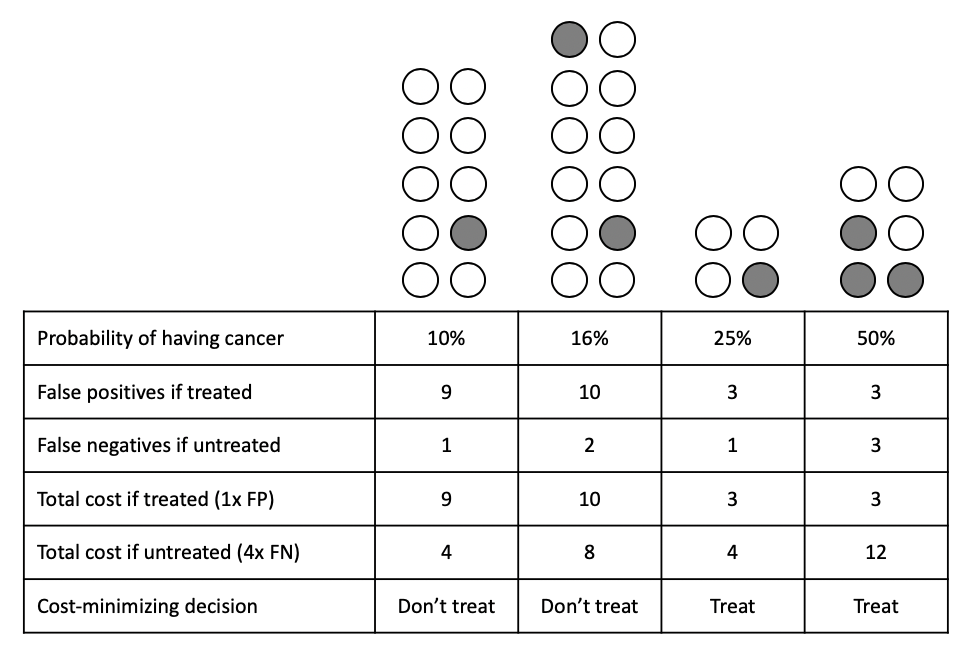}
         \caption{Male patients}
         \label{fig:cancer_risk_b}
     \end{subfigure}
  \caption{Hypothetical patients in a cancer treatment scenario. Dots symbolize patients at a given risk level, while solid dots denote patients who turn out to have cancer. Regardless of differences in the distribution of risk factors among the two different groups, the cost-minimizing treatment policy for both groups treats patients if and only if the probability they have cancer exceeds a certain threshold.}
  \label{fig:cancer_risk}
\end{figure*}

\subsection{Benefits and harms in binary decision making}

To see how this approach could be applied, consider a doctor deciding whether to prescribe chemotherapy to patients to treat potential cancers. The doctor cannot know for sure which patients have cancer, though they can divide the patients (perhaps using expert judgement, a diagnostic test, or a machine learning algorithm) into four categories of increasing risk, as shown in Fig.~\ref{fig:cancer_risk_a}. The probability of having cancer within each category is known, but there is no way for the doctor to identify exactly which patients within each category this probability will materialize for.

To a first approximation, there are two types of harms, or costs, to consider in this situation: the cost of a false positive, where the patient receives unnecessary and unpleasant chemotherapy; and the cost of a false negative, where the cancer goes untreated and may kill the patient.\footnote{One could equivalently describe avoiding these costs as benefits, though this change of reference point wouldn't change the analysis} There is a trade off between these costs---the aggressive prescription of chemotherapy will decrease false negatives but increase false positives, while a reluctance to pursue treatment will do the opposite. Resolving this tradeoff requires an assessment of, and judgement about, the \emph{relative} cost of false positives and false negatives. Suppose that, after consulting with subject matter experts and relevant stakeholders, the doctor determines that an undiagnosed cancer is 4 times worse than unnecessary chemotherapy. This \emph{cost ratio} captures what we mean by a system's tradeoffs---saying a false negative is four times more costly than a false positive is saying we would trade four false positives for one false negative, and vice versa. We note that in reality the choice to pursue a given medical treatment is also a highly personal decision that patients and their caretaking teams are involved in; we simplify the process here and assume that all patients have the same cost ratio for the purpose of illustration.

Fig.~\ref{fig:cancer_risk_a} shows the total cost of choosing to treat the patients in a given risk category (equal to the number of false positives this produces) compared to the total cost of choosing not to treat them (equal to four times the number of false negatives). Immediately, we can see the cost-minimizing strategy: treat only those patients in the two highest-risk categories. The optimal policy takes the form of a \emph{threshold}---we treat everyone above a certain level of risk and no one below it. 
We can now think about the general case of choosing thresholds when patients each have a continuous risk score $s$, and false negatives are $c$ times more costly than false positives (i.e.~the cost ratio is $c$). Let
\begin{equation}
    \textrm{cost}(t) = \textrm{FP}(t) + c\textrm{FN}(t)
    \label{eq:cost}
\end{equation}
describe the total cost of all decisions when a threshold $t$ is applied. We minimize the expected cost by taking the derivative with respect to $t$ and setting it to zero:
\begin{align}
    \frac{d\E[\textrm{cost}(t)]}{dt} &= \frac{d\E[\textrm{FP}(t)]}{dt} + c\frac{d\E[\textrm{FN}(t)]}{dt}\\
    0 &= -f(t^*)\E\left[1-Y|s=t^*\right] + cf(t^*)\E\left[Y|s=t^*\right]
\end{align}
where $f(s)$ is the density of cases with a given score $s$ and $\E[Y|s]$ is the rate of cancer cases among these patients. The derivative has a straightforward explanation: the decrease in false positives (or increase in false negatives) created by raising the threshold slightly is the product of the number of cases at the threshold and the fraction of those cases that were negative (or positive). Rearranging the equation gives us the optimality condition:
\begin{align}
    t_\textrm{impl}^*=\E\left[Y|s=t^*\right] &= \frac{1}{1+c}
    \label{eq:opt_threshold}
\end{align}
\emph{the cost-minimizing threshold is the score at which cases have a $\sfrac{1}{(1+c)}$ probability of being positive.}\footnote{This threshold is a global optima as long as $\E[Y|s]$ monotonically increases in $s$, and it is the unique optimum if $\E[Y|s]$ is strictly monotonic. In practice it doesn't matter if the optimal threshold is not unique---there will be a range of thresholds (the closed interval between $\sfrac{1}{6}$ and $\sfrac{1}{4}$ in the example in Fig.~\ref{fig:cancer_risk_a}) that all produce the same, cost-minimizing decisions.} We call the probability $\E\left[Y|s=t\right]$ the \emph{implied threshold}. Returning to our example we can now compute the precise cost-minimizing treatment strategy when $c=4$: treat any patient with a greater than 20\% chance of having cancer (i.e.~the optimal implied threshold is 20\%).

\subsection{Introducing subgroups}

Up until this point there has been no notion of subpopulations. Now imagine that the patients in Fig.~\ref{fig:cancer_risk_a} are female and the patients in Fig.~\ref{fig:cancer_risk_b} are male. This particular cancer affects a smaller fraction of male patients: 22\% of males have it, compared to 26\% of females. As a result, there are more males in the lower-risk categories and fewer in the higher-risk categories. We want ensure that our treatment strategy is fair to men and women.

We have established the optimal treatment approach for female patients: treat all patients with a >20\% chance of having cancer. But what is the optimal approach for male patients, who have a different base rate and distribution among the risk categories? Perhaps surprisingly, \emph{as long as the cost of decisions is the same} (i.e.~false negatives are four times more costly than false positives) the cost-minimizing approach is identical: we should treat all male patients with >20\% chance of having cancer. This is because Eq.~\ref{eq:opt_threshold} does not depend on $f(s)$, the distribution of risk among the group in question; nor does it depend on the base rate. 

This has important consequences for the study of fair machine learning. Consider an alternate approach that is popular in the fair machine learning literature: equalizing error rates~\cite{hardt2016equality}. One might argue that a false negative cancer diagnosis is so costly that fairness demands that our decisions produce equal false negative \emph{rates} for male and female patients. The cost-minimizing approach does not produce such equality---the false negative rate for male patients is 43\%, while the false negative rate for female patients is 22\%.

Equalizing these rates would require some combination of treating more male patients and treating fewer female patients. However, since we've already chosen the treatment strategy that minimizes costs faced by both male and female patients, such an intervention would inevitably make at least one group worse off without making the other group better off. The same is true for equalizing false positive rates, or treatment rates (as would be required by demographic parity). A demand that these fairness criteria be satisfied, then, must reflect a judgement that equality in a particular ratio (in our example, the number of false negatives divided by the total number of cancer cases) is of greater fairness value than the minimization of the actual harm caused by errors to members of both groups. There are many reasons why this would not be an appropriate judgement to make in practice. 

This is not to say that equal implied thresholds are always appropriate. Imagine if the cancer in question was more aggressive in female patients, such that a false negative was more likely to lead to death. In this case, the cost of a false negative (relative to the cost of a false positive) is greater for female patients than male patients, and the cost-minimizing threshold for female patients decreases. Our approach to fairness, which focuses on concrete impacts to people, would therefore require that different thresholds be applied to treatment decisions for male and female patients. Note that this treatment regime, which is cost-minimizing for both groups, actually increases the differences between false negative rates for male and female patients. This further illustrates the disconnect between such error rates and the well-being of decision subjects.

\subsection{Competing values}
\label{sec:competing_values_fairness_principle}
In cancer treatment, the potential cost of false positives \emph{and} false negatives fall on the same patient, making it easy to argue for the decision-making strategy that makes patients from all groups best off~\cite{ustun2019decoupled}.\footnote{Though, troublingly, some papers still advocate for decreased diagnostic performance in healthcare settings in the name of ``fairness''~\cite{pfohl2019creating}.} But what about cases where the tradeoff occurs \emph{between} groups? Consider the detection of spam on social media. The cost of a false positive principally falls on the publisher whose content is filtered (though consumers are also deprived of the opportunity to see this content), while the cost of a false negative falls on the consumer whose feed is contains low-quality content. As a result, the cost-minimizing approach for publishers would classify almost no content as spam,\footnote{Scrupulous publishers might argue for some spam enforcement so that their good content doesn't get drowned in a sea of spam, but a concern for false positives affecting their content would keep them from advocating for a system as strict as consumers would prefer.} while consumers would be better served by a more aggressive filtering system.

In these situations, a choice must be made about how the system will prioritize costs to different stakeholders. This prioritization doesn't have to reflect objective costs---cancer is objectively more costly than being caught in the rain, but it's perfectly acceptable for umbrella manufacturers to prioritize keeping people dry. Instead, it should reflect the goals and values of the system that have been carefully considered at the product and policy levels of fairness analysis. For example, we might decide that, for the spam filtering system, a false positive (affecting content producers) is ten times more costly than a false negative (affecting consumers). There is no single right answer here---different platforms make different tradeoffs in content moderation and different types of policy-violating content on the same platform might present significantly different costs---but being explicit helps ensure that the decision is made deliberately and consistently over time. The approach that achieves the desired prioritization is still described by Eq.~\ref{eq:opt_threshold}: if false positives are tens times as costly as false negatives ($c=0.1$), the optimal threshold is $\sfrac{1}{(1+c)}=0.91$. 

This is also true for any subset of the decisions. 
Consider the total costs created when potential spam published by a user from group $a$ may be seen be a consumer from group $b$:
\begin{equation}
    cost_{ab}(t_{ab}) = \textrm{FP}_{ab}(t_{ab}) + c_{ab}\textrm{FN}_{ab}(t_{ab})
\end{equation}
This takes the same form as Eq.~\ref{eq:cost}, and as a result the cost-minimizing threshold depends only on $c_{ab}$, the cost to consumers from group $b$ when they see spam from group $a$ relative to the cost to publishers from group $a$ when non-spam content is filtered out of a user from group $b$'s feed. This motivates the key fairness principle that drives our analyses of binary decision making:
\begin{quote}
\emph{If we believe that a false positive is equally costly whether it affects somebody in group A or group B, and that a false negative is equally costly whether it affects somebody in group A or group B, we should apply the same implied threshold to decisions affecting both groups. Doing so will minimize errors no matter who they affect.}  
\end{quote}

The antecedent won't always be true; for example, a platform might decide to explicitly prioritize female publishers by treating a false positive affecting female publishers as more costly than one affecting male publishers. In the criminal justice domain, system designers might opt to treat false positives as more costly for groups who, for example, tend to be penalized more harshly in future circumstances for having a history of incarceration. Again, these are decisions that are most appropriately made at the product and policy levels of fairness analysis. If such a prioritization was agreed upon, our implementation fairness approach could still be used to ensure the appropriate---and in this case, different---implied thresholds are being applied in practice.

\subsection{Analyzing existing systems}
\label{sec:implied}

Many real-world systems are not designed according to the three-step approach laid out in Section \ref{sec:an_approach}. Rather than making explicit choices about how to trade off between different costs and benefits, they arrive at the set of decisions they make through heuristics, inertia, and---potentially---mistakes. In such cases, we may be called to assess the fairness of a system without knowing exactly how it was designed. Our approach can still be used in these circumstances, it just needs to be reversed. Instead of choosing a set of tradeoffs and then ensuring that they are achieved in practice, we can work backwards from the decisions the system is currently making to determine the set of tradeoffs (i.e.~cost ratios) that make these decisions cost-minimizing. We call these the \emph{implied} tradeoffs of a system.\footnote{Welfare economics and optimal taxation theory use a similar concept---``inverse welfare weights''---to describe how a decision maker must weigh different individuals' welfare make a set of decisions welfare-maximizing in the aggregate~\cite{kasy2020fairness,saez2016welfare}.} In the case where the relevant costs are due to false positives and false negatives, we can invert Eq.~\ref{eq:opt_threshold} and use the implied threshold determine the cost ratio the decision maker is operating under:
\begin{equation}
    c = \frac{1-E[Y|s=t]}{E[Y|s=t]}.
    \label{eq:implied_cost}
\end{equation}

Consider a treatment regime for the patients in Fig.~\ref{fig:cancer_risk} where female patients in the two highest-risk categories and male patients in the \emph{three} highest-risk categories receive chemotherapy. Assuming this treatment regime minimizes some conception of cost for male and female patients, we can apply Eq.~\ref{eq:opt_threshold} to compute the implied cost ratios for both groups. $\E[Y|s=t, \textrm{female}]=0.25$, since 25\% of the patients at the threshold of treatment (the second-riskiest group) actually have cancer. This implies a female cost ratio of 3. In comparison, $\E[Y|s=t, \textrm{male}]=0.16$, implying a male cost ratio of 5. This treatment regime implies we believe that false negatives are more costly for male patients than female patients (or, equivalently, that false positives are more costly for female patients). If this reflects our considered belief, then it is fine. If this was not intended, and instead occurred because of some error or oversight (for example, the systematic overestimation of male patients' cancer risk), then we can consider ways to fix this. Section \ref{sec:implementation_fairness} describes how we measure implied tradeoffs in algorithmic decision making and human labeling.

\section{Measuring fairness in practice}
\label{sec:implementation_fairness}

To illustrate the principles behind our measurement approach, we'll use the example of a system designed to identify and remove posts that violate an online platform's policy against bullying and harassment. The system combines decisions made by humans with binary decisions made on the basis of machine-learned predictions. Human labelers are employed to determine whether a given post violates the policy; these labels then inform both immediate enforcement (i.e.~the removal of posts from the platform) and the training of the machine learning system. The machine learning system, in turn, produces predictions that are used both to triage potential harassment for human review and to automatically remove the most obvious cases of bullying.

This example clearly illustrates the breadth of questions that must be answered before a technical fairness analysis is even attempted, and the utility of distinguishing among product, policy, and implementation fairness. For example, how should the policy define bullying and harassment? Do all expressions of harassment qualify, or does it depend on the subject of that bullying's membership in certain demographic groups? If the latter, which groups are protected? What is the nature and degree of harm potentially caused by false positives and false negatives, and how should the system trade off between those errors in enforcement? These are incredibly challenging questions (as evidenced by the fact that different internet platforms, social institutions, and liberal democracies take different approaches to objectionable speech) which we will not attempt to answer in this paper. Still, it is important to note that the empirical questions we consider at the implementation fairness level are only a small part of the fairness puzzle.

In this section we first apply our approach to decisions made by the machine learning system---the familiar problem of fairness in supervised learning. We then show that the same conceptual approach can be used to study the fairness of human decisions with respect to a ground truth.

\subsection{Model fairness}
\label{sec:algorithmic_decisions}

The simplest and most ubiquitous type of algorithmic decision is a binary decision based on the prediction of a machine learning model---one action is taken if the probability of the predicted event is suitably high, another action taken otherwise. A system designed to automatically (without human intervention) identify and remove violating bullying and harassment takes this form.\footnote{Some might argue that such automated intervention would not be appropriate; such a conversation would be situated at the \emph{product} level of our fairness analysis while investigations of the comparative performance of such a system would be a question of \emph{implementation}.}

Consider the task of measuring the fairness of such a system. First, it is necessary to decide which subpopulations to compare. We could choose to measure the system for different groups of content consumers or producers, different groups targeted by bullying (which may not be the same as the intended consumer), or intersections of these groups. For the sake of exposition, though, let's consider two groups of content producers: groups ``A'' and ``B''. Differences in the base rate of outcomes are both extremely common in situations where fairness concerns are present (including in content moderation), and give rise to important impossibility results in the fair machine learning literature \cite{kleinberg2016inherent}. In our hypothetical scenario, group A's content is more likely to be bullying or harassment than group B's content.

Following the approach outlined in Section \ref{sec:an_approach}, we then have to identify the costs and benefits of the decision in question, and how these may be in tension for different groups. As in our previous examples, the most salient costs are the costs of false positives (erroneous removal of content) and false negatives (failing to remove bullying or harassment). There is no obvious tension between the costs experienced by \emph{producers} in each group---taking a more lenient approach to content from group A, for example, does not require taking a more aggressive (or more lenient) approach to content from group B. Instead, the relevant tension is between the welfare of content producers and content consumers. To navigate this tension, we would need to be explicit about how we prioritize the removal of violating content from each group. For now, let's say we have decided that the appropriate cost ratio is the same for both groups: though group A produces more bullying content, an individual instance of bullying is no more costly coming from a user in group A than coming from a user in group B (we return to this decision in the next section). Similarly, a false positive is equally bad whether it affects users from group A or group B.  As a result, our fairness principle in Section \ref{sec:competing_values_fairness_principle} holds, and we should expect to see equal thresholds being applied to content from both groups.

To ensure the desired tradeoff (equal cost ratios) is being achieved in practice, we must ensure the implied threshold is equal for content produced by both groups. It's important to distinguish between two different thresholds here. The \emph{decision threshold} $t$ defines how the output of the model (the score $s$) is mapped to decisions $\hat{Y}$ (i.e.~whether the post is removed):
\begin{equation}
    \hat{Y}_i= \textbf{1}\{s_i\ge t\},
\end{equation}
whereas the \emph{implied threshold} is the probability of the outcome corresponding to the decision threshold:
\begin{equation}
    t_\textrm{impl}=\E[Y|s=t].
\end{equation}
If the score $s$ is \emph{calibrated}, such that $s=\E[Y|s=s]$, then the decision threshold and implied threshold are identical. In general, though, we cannot assume the model being assessed produces calibrated predictions (some models, like support vector machines, don't even produce scores in $[0,1]$).

There are a number of challenges that must be overcome to estimate the implied threshold from the output of the actual machine learning system. First, when the scores are continuous the score will almost never take on exactly the threshold value. As a result, we could approximate the implied threshold by conditioning on the score taking some value within a window around the threshold:
\begin{align}
    \E[Y|s=t] \approx& \E[Y|t_l\le s \le t_u] \quad \textrm{where} \\
    &t_l< t < t_u. \nonumber
\end{align}
Unfortunately, this approximation is often poor because of how scores are typically distributed. Bullying and harassment makes up a very small fraction of all posts on content platforms, so lower scores would generally be more common than higher scores, and samples in the window would disproportionately come from the lower end of the window. Therefore, simply averaging the outcomes of posts with scores within the window will tend to underestimate the implied threshold.

To address these problems, we first fit a weighted linear regression to posts within a symmetric region (of half-width $d$) around the threshold using tricubic weights:
\begin{align}
Y_i &= \beta_0 + \beta_1(s_i-t) + \epsilon_i \\
    w_i &=
    \begin{cases}
        \left[1 - \left(\frac{|s_i - t|}{d}\right)^3\right]^3 & s_i \in [t - d, t + d] \\
        0 & \text{otherwise}
    \end{cases}.
\end{align}
Then, the intercept $\beta_0$ is our estimate of the implied threshold. We call this estimator the \emph{prevalence at the threshold}. Statistically significant differences in this value between groups A and B would indicate that our system is not minimizing errors regardless of which group they affect, and is instead prioritizing errors affecting one group.

\subsubsection{Alternative approaches}
\label{sec:alternatives}
It is worth revisiting why we have not adopted alternative approaches to fair machine learning popular in the computer science literature. It's clear that statistical parity isn't appropriate: removing an equal fraction of posts from all groups, regardless of how much bullying they engage in, is not a viable approach for a real-world content moderation system. But the reason why we haven't chosen to, as a baseline, equalize error rates (either the false positive rate, the false negative rate, or both) is more subtle.

Existing research has established that the cost-minimizing approach to equalizing error rates uses different implied thresholds for each group~\cite{corbett2017algorithmic}. In practice, the direction that thresholds must be adjusted is determined by a group's prevalence: equalizing error rates means applying a higher (more lenient) threshold to the group producing more bullying and harassment content (group A), and a lower (stricter) threshold to the group producing less of such content content (group B).\footnote{This is because, for most score distributions and thresholds, applying the same implied threshold to both groups will lead to a higher false positive rate and lower false negative rate for the higher-prevalence group. It is possible to construct score distributions where this does not occur (in which case equalizing error rates would require stricter thresholds for the higher-prevalence group), but these tend to be multi-modal in a way that we have not observed in practice.} Since we know that the implied threshold and the cost ratio are related by Eq.~\ref{eq:opt_threshold}, such differences in thresholds can only be rationalized by a decision to treat false negatives from group A (the higher-prevalence group) as \emph{less} costly than the same errors affecting group B. In other words, equalizing error rates would imply that we believe that bullying and harassment is not as bad when produced by the higher-prevalence group.

It is important to note that there may be cases where false positives or false negatives \emph{do} have different costs for content producers in group A or B, or for content consumers in group A or B; for example, if violating content produced by group B has a higher risk of leading to serious physical or psychological harms. In such cases, it may be deemed appropriate to apply different thresholds---but we hold that decisions about whether to treat subgroups differently as a fairness remedy should be made explicitly, with subject matter experts, and aligns with the \emph{policy} or \emph{product} dimension of fairness, rather than an unintended outcome of opting for statistical parity or equalized error rates within the \emph{implementation} dimension. 

Finally, error rates can depend on ``easy'' decisions that are irrelevant to the question of fairness. Most posts are obviously not bullying or harassment (e.g.~``Happy birthday!''), and will never be removed by any content moderation system. And yet, since the number of non-violating posts makes up the denominator of the false positive rate, a group's false positive rate is affected by the number of obviously benign messages they post. It is clearly undesirable for a fairness assessment of a content moderation system to be affected by a group's tendency to share benign messages---but that would be the implication of considering false positive rates.

This problem is further exacerbated by the presence of adversarial behavior. Imagine bad actors from some group realizing that the system was designed to equalize false negative rates. They could spam the system with easy-to-identify instances of harassment, driving down their group's false negative rate. The system would be forced to respond by applying a more lenient threshold to content from the offending group, increasing the false negative rate to compensate and maintain error rate parity. By flooding the platform with obvious harassment, the bad actors would have forced us to be more lenient to the rest of their posts! This set of incentives would be problematic for a large-scale content moderation system to adopt. We note that as researchers continue to iterate on implementation fairness definitions and approaches, innovations in fair ML research may yet inform adaptations to our approach in the future.

\subsection{Label fairness}
\label{sec:label_bias}
As with most approaches to fair supervised learning, the approach described in the previous section assumes the outcome being predicted ($Y$) is measured accurately in the data used to assess the system. There are some cases where this assumption is reasonable---for example, websites can perfectly measure whether users click on a given button. However, in many cases, such as identifying bullying, the labels themselves are generated through human judgement, and may thus embed human biases. This is of concern for at least three reasons. First, accurate labels are needed to compute most model fairness metrics, including the metric in Section \ref{sec:algorithmic_decisions}. Second, supervised learning systems trained on biased labels will learn those biases. Finally, labelers' decisions might be used to directly intervene in the system.
In this section, we describe how our high level fairness approach can be applied to assess human decision making, in the case where decisions can be compared to a ground truth.

In our bullying and harassment example, the decision being made by human labelers is whether a given post violates a bullying policy as written. These decisions won't always be correct---labelers may misunderstand the policy or the post, make a mistake, or be misled by implicit or explicit biases. To track these errors, we also collect (for a subset of posts) the judgement of an expert in applying the written policy, whose decisions provide the ground truth for each post. A fairness measurement dataset, then, would consist of a set of tuples $(Y_{ij}, Y_i^*)$ for every label, where $Y_{ij}$ is the label provided by labeler $j$ to post $i$, and $Y_i^*$ is the expert-provided ground truth for that post.

As before, we can summarize the costs created by the human labeling process in terms of false positives and false negatives with respect to the ground truth. Ideally, then, we'd proceed as we would in the algorithmic decision making case: determine the implied threshold that labelers are using for posts from each group, and ensure it reflects the appropriate cost ratio. However, while in the algorithmic case we knew the decision threshold and which posts had scores close to the threshold, in the human decision making case we are missing all of this information. With only a binary label and a binary ground truth for each post, one might be tempted to abandon efforts to estimate the implied threshold and instead return to comparing something like group error rates. But this would be a mistake for the same reasons as described in Section \ref{sec:alternatives}: error rates are driven by easy decisions, both obviously benign (e.g.~``Happy birthday!'') and obviously violating. These are trivially easy for a human to judge correctly, and because of this are uninformative about a labeler's possible bias. We need a means of estimating the implied threshold with just a set of label/ground-truth pairs.

\subsection{Signal Detection Theory}

A promising approach comes in a model of human decision making developed by psychologists: Signal Detection Theory (SDT)~\cite{green1966signal}. Applied to the labeling of bullying or harassment, SDT models human decision making as follows: upon seeing a post, the labeler mentally accumulates evidence for and against the proposition that the post is policy-violating. The result of this accumulation is called the \emph{signal}. The labeler also conceives of a threshold (sometimes called the \emph{criterion}); when the signal exceeds the threshold they report the post as violating, when it doesn't they report the post as benign.\footnote{ It's important to remember that SDT is a \emph{model} of behavior---we're not suggesting that content labelers could report an actual numerical threshold if asked.}

\begin{figure}[t]
    \centering
    \includegraphics[width=0.8\columnwidth, trim=5 70 5 0]{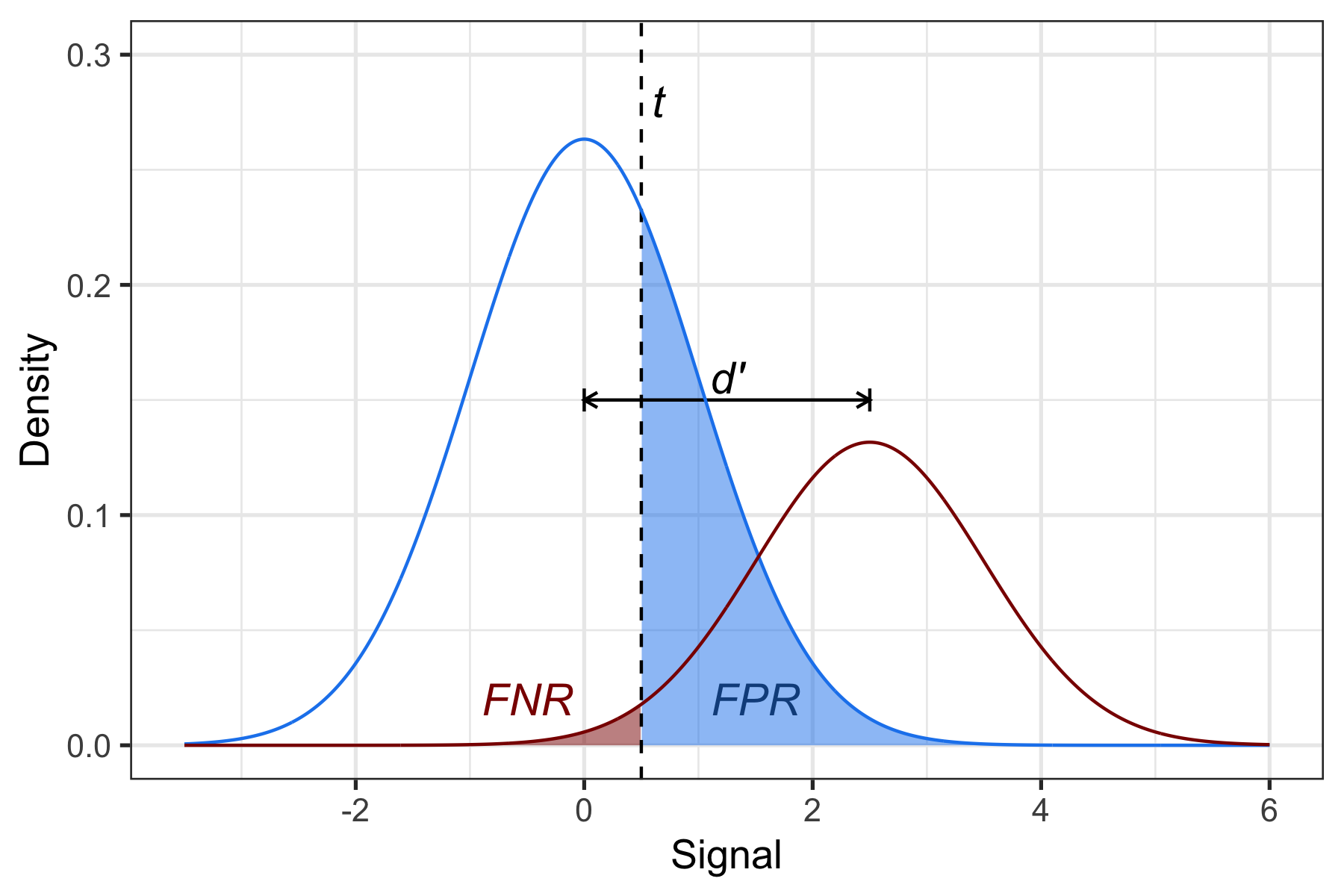}
    \caption{The signal detection theory model of decision making. Negative examples produce signals distributed according to the standard normal (blue curve). Positive examples produce signals sampled from a normal distribution with mean $d'$ and unit variance (red curve). A threshold $t$ is applied to turn signals into decisions. As a result, the false positive rate is simply the fraction of the negative distribution above the threshold (shaded blue), while the false negative rate is the fraction of the positive curve below the threshold (shaded red). Given observed false positive and false negative rates, we can therefore compute the $t$ and $d'$ values required to generate these error rates.}
    \label{fig:sdt}
\end{figure}

Statistically, SDT models the distribution of the signal as a mixture of Gaussians: one Gaussian for benign posts and another for violating posts. Figure \ref{fig:sdt} illustrates the SDT model. Signal detection theory has been used to study human decision making in many different contexts since its development in the 1940s and `50s, including: military radar signals~\cite{green1966signal}, medical diagnosis~\cite{stewart2004detection}, child welfare decisions~\cite{mumpower2014signal}, and policing~\cite{pierson2018fast}. 

Since the scale of the decision variable is arbitrary, we can (without loss of generality) choose the distribution for benign posts to be the standard normal. By further assuming that the Gaussians have equal variance,\footnote{Having equal variances ensures that the probability of a post being violating is monotonically increasing in the signal (see \citet{pierson2018fast} for proof and further details)}
SDT defines a two parameter family of signal distributions parameterized by the \emph{prevalence} ($\phi$) and the \emph{separation} ($d'$, pronounced ``dee-prime''). Prevalence is the fraction of posts that are violating, and therefore defines the mixing proportions of the Gaussians. Separation is the distance between the Gaussians' means. When separation is high there is little overlap between the signals produced by violating and benign posts, making it easy for labelers to correctly distinguish between them. When separation is low the distributions have substantial overlap, and labelers make more mistakes.\footnote{In fact, separation is directly related to AUC, the probability that a randomly chosen violating post will have a higher signal than a randomly chosen benign post: $d' = \sqrt{2}\Phi^{-1}(\textrm{AUC})$ (where $\Phi^{-1}(\cdot)$ is the inverse cumulative distribution function for the standard normal).}

Figure \ref{fig:sdt} shows how, for a given separation and threshold, the signal detection theory model implies a certain false positive rate and false negative rate. In particular:
\begin{align*}
    \textrm{FPR}(d', t) &= 1-\Phi(t)\\
    \textrm{FNR}(d', t) &= \Phi(t-d'),
\end{align*}
where $\Phi(\cdot)$ is the cumulative distribution function for the standard normal. Therefore, we can use the observed error rates to infer the model parameters:
\begin{align*}
    t &= \Phi^{-1}(1-\textrm{FPR}) \\
    d' &= t-\Phi^{-1}(\textrm{FNR}).
\end{align*}
Note that the threshold $t$ is not the implied threshold, since it is defined in signal space. But SDT allows us to compute the implied threshold using the prevalence and Bayes' rule:
\begin{align*}
    E[Y|s=t] &= P(Y=1|s=t)\\
    &= \frac{P(Y=1)P(s=t|Y=1)}{P(Y=1)P(s=t|Y=1) + P(Y=0)P(s=t|Y=0)}\\
    &= \frac{\phi \textrm{N}(t-d')}{\phi \textrm{N}(t-d') + (1-\phi) \textrm{N}(t)}\\
    &= \frac{1}{1+\frac{1-\phi}{\phi}\exp\left(-td'+d'^2/2\right)}
\end{align*}
($\textrm{N}(\cdot)$ is the density function of the standard normal distribution.)
Finally, we can plug the implied threshold into Eq.~\ref{eq:implied_cost} to recover the cost ratio that explains labelers' decisions:
\begin{align}
    c &= \frac{1-\phi}{\phi}\exp\left(-td'+d'^2/2\right).
\end{align}
Comparing cost ratios (or, equivalently, implied thresholds) between groups allows us to determine whether labelers are making appropriate judgements about content from different groups. In particular, if we believe labeling errors are equally costly regardless of the group they affect, we should ensure labelers are acting accordingly by applying the same implied threshold to all groups.

\subsubsection{Limitations of the approach}

While the two-parameter mixture model is flexible~\cite{pierson2018fast}, it cannot capture all plausible signal distributions. For example, imagine there were three distinct types of posts: those that are obviously violating, those that are obviously benign, and those that are genuinely ambiguous. Signal detection theory cannot model such a tri-modal distribution, so it will produce incorrect estimates of the implied threshold in this case. \citet{pierson2018fast} attempt to address this by allowing the SDT parameters to vary according to decision covariates, but this requires a substantially more complex Bayesian inference procedure. We elect to use the simpler model to make it easier to scale the approach to many different labeling tasks.

An important direction for future work on labeling fairness centers on the separation parameter. A low $d'$ for some set of decisions means that labelers have trouble distinguishing violating posts from benign posts. Currently, however, it is difficult to determine whether this trouble is due to the labeling problem being fundamentally hard in some sense, or because labelers are making mistakes---unconsciously or otherwise---in a way that may be attributable to their own bias (or other factors). For example, a labeler's bias against certain posts might manifest not in them erring towards false positives or false negatives (which SDT can measure using the cost ratio), but in them being indifferent about making errors in general, leading them to rush their decisions. In principle, this type of bias is still amenable to being measured with an extension of our tradeoff-focused approach---now the labeler is trading off between the cost of a false positive, the cost of a false negative, \emph{and} the cost of their labeling time. The drift-diffusion model is an extension of SDT that attempts to account for decision time in this way~\cite{krajbich2011drift}.

\section{Practical challenges and open questions}
\label{sec:practical_challenges}
\subsection{Mitigating implementation fairness issues}
Many machine learning papers that propose new model fairness metrics also develop algorithms to satisfy these metrics, either through optimization constraints~\cite{corbett2017algorithmic, zafar2017disparate,agarwal2018reductions} or by incorporating the metric into the training loss function~\cite{kamishima2011fairness}. These approaches purport to automatically ensure that a new model is ``fair'', but each necessarily reduces the performance of the model, increasing the number of people affected by model errors. We believe such approaches are often unwise: measured unfairness is a symptom of deeper problems in a system that likely can't be solved through a tweak in the optimization process. Furthermore, designing the optimization process such that fairness issues are never measured risks papering over these problems while often making decision subjects worse off. 

Unfairness in a model has many different possible causes, including: a lack of training data, a lack of  features, a misspecified target variable, or measurement error in the input features. None of these problems are amenable to typical machine learning optimization---their solutions exist outside the bounds of the optimization problem. The challenging upshot of this is that there is no silver bullet for mitigating implementation fairness issues. Instead, we believe that the measurement of fairness issues should prompt a deep dive into the model to diagnose and remedy the root cause of the issue. 

This is especially true when trying to mitigate label bias concerns, since this always means changing human behavior. Fortunately, psychologists have demonstrated that labelers' will change the thresholds they apply when incentivized~\cite{curran2007criterion}. If the problem can be isolated to specific labelers who are being too strict or lenient, they could be be nudged into applying a more appropriate thresholds. If the problem is systematic, one should investigate the labelers' guidelines, how labelers are selected, and whether they are appropriately representative.\vspace{-0.2cm}

\subsection{Group characteristic data}

To measure potential bias affecting a sensitive subpopulations, one generally needs to know which people affected by the system are members of that group.
However, the sensitive nature of the characteristics most relevant to fairness---including gender, ethnicity, religion, and national origin---poses important challenges for efforts to understand fairness in practice~\cite{bogen2020awareness}.

First, the entity trying to study fairness may lack subgroup information. While internet platforms may solicit a user's age and gender, they rarely collect or infer information about a user's race, ethnicity, religion, or sexual orientation. Collecting or otherwise obtaining this data raises privacy, ethical, and representational questions. In some cases, unresolved tensions between privacy and fairness have meant that we have lacked the data to perform fairness analyses pertaining to certain subgroups.

Even when the data is available, our methodologies require creating discrete groups out of complex identities. Discretizing someone's gender, age, or race will necessarily lack important details about their lived experience, or worse, may re-enforce historical categories that fuel discrimination or erase identities. But statistical measurement requires grouping users \emph{somehow}. Fairness practitioners need to make sure that (a) group definitions and data are as reflective as possible of users' self-identities, (b) group designations are sufficiently flexible to capture a wide range of fairness concerns, and (c) users are provided sufficient control to fix mistakes in groupings.

Finally, subgroup information will always be subject to measurement error; even for self-reported attributes like gender users might decline to specify, choose an option at random, or make a data-entry mistake. Furthermore, some practitioners use \emph{inferred} sensitive characteristics for fairness measurements and interventions~\cite{geyik2019fairness}. Such inferences are likely to increase the number of errors in subgroup assignment by orders of magnitude. An open question remains as to how these errors could affect fairness measurements, especially if these subgroup identification errors are correlated with decision-making errors.\vspace{-0.1cm}

\subsection{Complexity of systems}
Using metrics that are correctly tailored to the potential benefits and harms that users may experience is central to our fairness approach. We have discussed metric recommendations for binary decision-making, but best practices do not yet exist to measure fairness for more complex model or system types. 

For example, models are often combined or have feedback loops. In these systems, measuring only individual components may not reveal issues that emerge only in their interactions, and conversely, individual component measurements may not necessarily support drawing conclusions about an overall system. Conducting measurement on each single model is a necessary starting point, but further research into how components may combine to create---or  reduce---fairness risk is needed.\vspace{-0.2cm}

\section{Conclusion}

This paper has presented an approach to addressing fairness challenges developed within the context of a large technology company. Our approach considers fairness at three levels---product, policy, and implementation---allowing us to direct analyses and interventions towards the appropriate part of the system, and to separate normative questions from statistical ones where appropriate. At the implementation level, we also presented a high-level approach to studying fairness questions grounded in the costs and benefits produced by decisions. Finally, we discussed that approach in two archetypal binary decision-making contexts: algorithmic decision making and human labeling.

\bibliographystyle{ACM-Reference-Format}
\bibliography{refs}

\appendix

\end{document}